\documentclass[journal]{IEEEtran}
\usepackage[latin9]{inputenc}
\usepackage{array}
\usepackage{multirow}
\usepackage{amsmath}
\usepackage{graphicx}

\makeatletter

\providecommand{\tabularnewline}{\\}

\usepackage{epsfig}
\usepackage{amssymb}
\usepackage{multirow}

\usepackage{cite}

\usepackage{graphicx}
\ifCLASSINFOpdf
\else
\fi
\usepackage{amsmath}
\usepackage{algorithmic}

\usepackage{array}
\usepackage{url}

\makeatother

\begin{document}

\title{Scene Recognition with Prototype-agnostic Scene Layout}

\author{Gongwei~Chen, Xinhang~Song, Haitao~Zeng, and Shuqiang Jiang,~\IEEEmembership{Senior~Member,~IEEE}\thanks{Corresponding author: Shuqiang Jiang}\thanks{G. Chen, X. Song, S. Jiang are with the Key Laboratory of Intelligent
Information Processing of Chinese Academy of Sciences (CAS) Institute
of Computer Technology, CAS, Beijing, 100190, China, are also with
University of Chinese Academy of Sciences, Beijing 100049, China (e-mail:
gongwei.chen@vipl.ict.ac.cn, xinhang.song@vipl.ict.ac.cn, sqjiang@ict.ac.cn).}\thanks{H. Zeng is with China University of Mining and Technology, Beijing,
100083, China, and also an intern with the Key Laboratory of Intelligent
Information Processing, Institute of Computing Technology, Chinese
Academy of Sciences, Beijing, 100190, China (email: haitao.zeng@vipl.ict.ac.cn).}}
\maketitle
\begin{abstract}
Exploiting the spatial structure in scene images is a key research
direction for scene recognition. Due to the large intra-class structural
diversity, building and modeling flexible structural layout to adapt
various image characteristics is a challenge. Existing structural
modeling methods in scene recognition either focus on predefined grids
or rely on learned prototypes, which all have limited representative
ability. In this paper, we propose Prototype-agnostic Scene Layout
(PaSL) construction method to build the spatial structure for each
image without conforming to any prototype. Our PaSL can flexibly capture
the diverse spatial characteristic of scene images and have considerable
generalization capability. Given a PaSL, we build Layout Graph Network
(LGN) where regions in PaSL are defined as nodes and two kinds of
independent relations between regions are encoded as edges. The LGN
aims to incorporate two topological structures (formed in spatial
and semantic similarity dimensions) into image representations through
graph convolution. Extensive experiments show that our approach achieves
state-of-the-art results on widely recognized MIT67 and SUN397 datasets
without multi-model or multi-scale fusion. Moreover, we also conduct
the experiments on one of the largest scale datasets, Places365. The
results demonstrate the proposed method can be well generalized and
obtains competitive performance.
\end{abstract}


\section{Introduction}

\begin{figure}
\noindent \begin{centering}
\includegraphics[width=1\columnwidth]{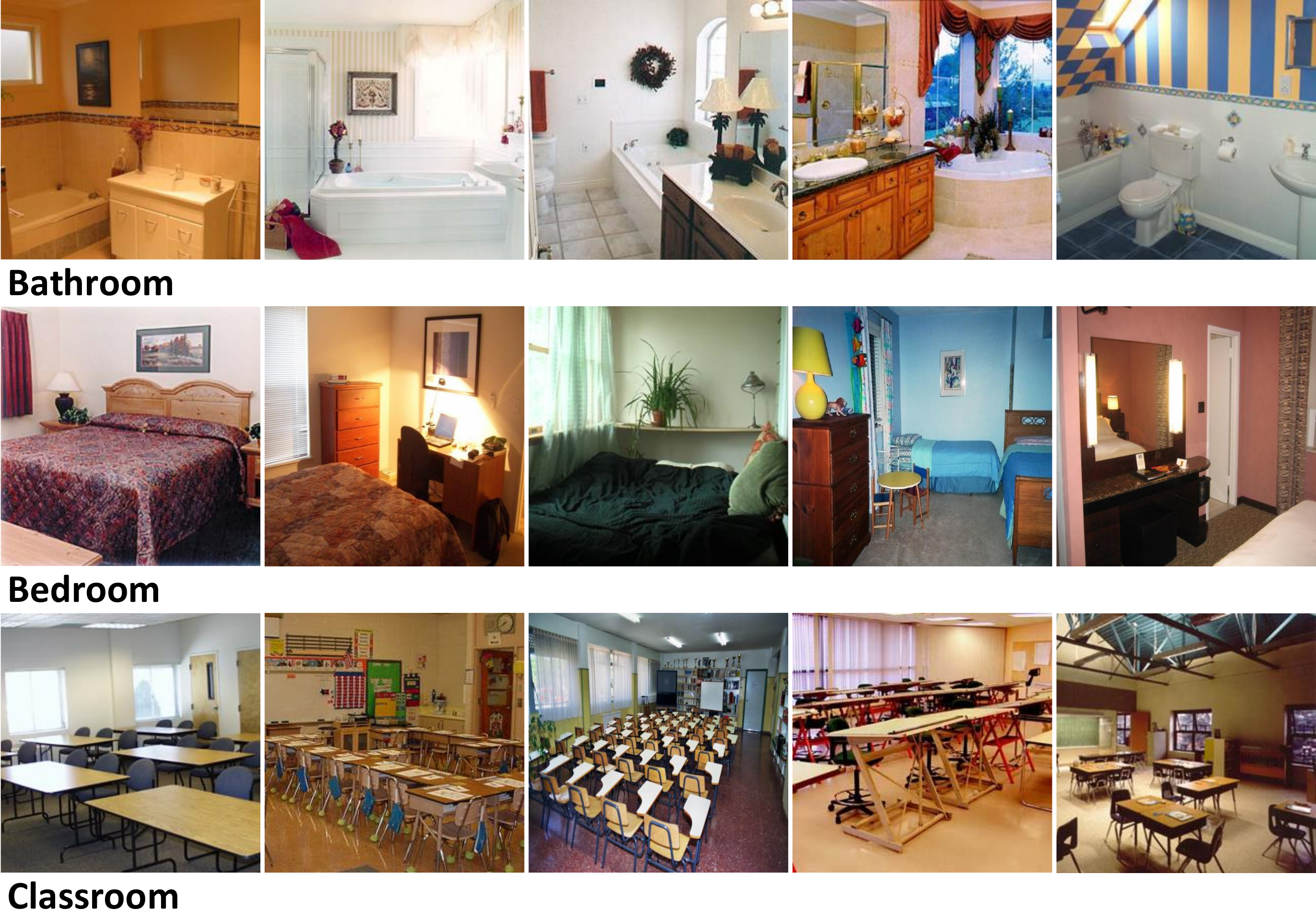}
\par\end{centering}
\caption{\label{fig:Intro-examples}Image examples from MIT67 datasets. Here
shows the images from three scene categories (``bathroom'', ``bedroom'',
and ``classroom''). It can be seen that the objects in each scene
can vary greatly in size and location, like ``bathtub'' in scene
``bathroom'', ``bed'' in scene ``bedroom'', ``desk'' in scene
``classroom''.}
\end{figure}

Scene images (e.g., ``classroom'', ``bedroom'') are usually composed
of specific semantic regions (e.g., ``desk'', ``bed'') distributed
in certain spatial structures. Exploring the local regions and their
spatial structures has been a long-standing research direction and
plays a crucial role in scene recognition \cite{MIT67CVPR2009,Izadinia2014CVPR,HRNN2016TIP}.
Due to the size and location changes of semantic regions (see Fig.\ \ref{fig:Intro-examples}),
the spatial structures of images have great diversity, which makes
it very difficult to represent them so as to adapt various image characteristics.
Thus, how to build and model such structural layout into image representations
is an obstacle problem.

Most existing methods \cite{MetaObject2015ICCV,HRNN2016TIP,AdaDisc2018ACMMMM,Liu2018AAAI}
model spatial structural information based on predefined grid regions
or densely sampled regions. These regions are in fixed sizes located
in a grid, forming a simple and constant structure as a common prototype
for all images, which results in rigid layout even with the extension
of multi-scale setting. Some earlier works \cite{MIT67CVPR2009,Pandey2011ICCV,Izadinia2014CVPR}
have attempted to learn several prototypes for each category images
with different models, such as constellation model in \cite{MIT67CVPR2009},
Deformable Part-based Model (DPM) in \cite{Pandey2011ICCV} and DPM's
variant in \cite{Izadinia2014CVPR}. These prototypes can be regarded
as templates with fixed topological structures for each scene category,
where the geometric relations of the components are obtained through
statistic learning. The spatial structure of each image is constructed
by conforming to the prototypes. Although more than one prototype
is usually used to characterize one scene category, such limited variety
is not comprehensive enough to cover the large intra-class structural
diversity of scene images. In contrast, our motivation is to design
a layout modeling framework to flexibly capture the unconstrained
spatial structures and effectively obtain discriminative patterns
from them.

In this paper, we propose Prototype-agnostic Scene Layout (PaSL) construction
method, which builds spatial structure for each single image without
conforming to any prototype. Given an image, PaSL is constructed with
the locations and sizes of discriminative semantic regions, which
are detected by only using the convolutional activation maps of this
image. Thus, PaSLs will vary from image to image and can flexibly
express different spatial characteristics of the images. Considering
the natural property of the graph to preserve diverse and free topological
structures, we frame the structural modeling process as a graph representation
learning problem. More specifically, we propose Layout Graph Network
(LGN) where regions in PaSL are defined as nodes and two kinds of
relations between nodes are encoded as edges. Through the graph convolution
\cite{GCN2017ICLR} and mapping operations of LGN, the topological
structure and region representations can be transformed into a discriminative
image representation.

The main idea of PaSL construction method is inspired by the ability
of pretrained CNNs to localize the meaningful semantic parts \cite{NetDissec2017CVPR}.
We make use of the convolution activation maps extracted from pretrained
CNNs to detect semantic regions, and aggregate them to generate discriminative
regions and form PaSL in an unsupervised way. The advantages of our
method is two folds. One is that the whole process is performed on
each image independently and can be easily extended to large scale
datasets. Another is that PaSLs derived from different pretrained
CNNs can yield comparable performances with same LGN, which demonstrates
they have considerable generalization ability. Besides constructing
PaSL, modeling it in graph structure is also an important contribution
in this paper. Conventional structural models in scene recognition
either have difficulty of optimization \cite{Pandey2011ICCV,Izadinia2014CVPR}
on large scale datasets or simplify the structural information \cite{HRNN2016TIP,Song2017TIP}.
In contrast, we build Layout Graph Network upon PaSL by reorganizing
it as a layout graph containing two subgraphs. These two subgraphs
aim to capture different kinds of relations, spatial and semantic
similarity relations between regions, respectively. Thanks to the
independence between these two kinds of relations, we can explore
structural information in a higher order space and easily encode it
into more discriminative features. Furthermore, the application of
graph convolution makes our model effectively handle various topological
structures and easy to be optimized with large amounts of data.

We evaluate our model on three widely recognized scene datasets, MIT67
\cite{MIT67CVPR2009}, SUN397 \cite{SUN397CVPR2010}, and Places365
\cite{Places2TPAMI2018}. The ablation study shows that our method
obtains up to $5\%$ improvements over baselines that neglect structural
information. Compared to current works on MIT67 and SUN397 that benefit
from multi-model or multi-scale fusion methods, our model even outperforms
them and obtains state-of-the-art results with single model in single
scale. When extending our model to one of the largest scale dataset,
Places365, it still shows competitive performance.

\section{Related Work}

\subsection{Scene Recognition}

In early works, bag-of-feature methods (like VLAD \cite{VLAD2010CVPR},
Fisher Vector \cite{FV2010ECCV}) with handcrafted features (like
SIFT \cite{SIFT2004IJCV}) have demonstrated great power on scene
recognition. However, these methods incorporate local information
in an orderless way, which loses spatial dependencies between local
regions. Then, some works further explore the spatial contextual dependencies
within bag-of-feature methods. Lazebnik \textit{et al.} \cite{Lazebnik2006CVPR}
proposed Spatial Pyramid Matching to exploit approximate spatial information
in a predefined grid. Parizi \textit{et al.} \cite{RBOW2012CVPR}
used a reconfigurable model operated on a grid to capture the spatial
information among regions.

Beyond this simple and fixed spatial information based on grids, some
works explore the complex and flexible spatial structures formed by
scene components in different ways. These works \cite{MIT67CVPR2009,Pandey2011ICCV,Izadinia2014CVPR}
construct the scene structures in a similar way to Deformable Parts
Model (DPM) \cite{Pandey2011ICCV}, DPM's variant \cite{Izadinia2014CVPR},
or a constellation model \cite{MIT67CVPR2009}. Based on these structural
models, a fixed number of structures for each scene category, which
can be named as scene prototypes, are learned. Then the spatial structural
information of each image is discovered by conforming to the default
structure of the most matched scene prototype. The spatial structures
derived from scene prototypes have difficulty covering the intra-class
variety of scenes. Differently, our approach can model the structural
layout for each image following its own characteristics.

Recently, Deep Learning methods, especially Convolution Neural Networks,
have been widely used in scene recognition. Some works \cite{Xie2017TCSVT,MFAFVNet2017ICCV,Liu2018AAAI,Chen2018PR}
combine bag-of-feature methods (like VLAD, Fisher Vector) or dictionary-based
method with CNN to explore discriminative local information in an
orderless way. To model spatial contextual dependencies, the works
\cite{HRNN2016TIP,Song2017TIP} learn a sequential model (like LSTM
\cite{HRNN2016TIP}) or a graphical model (MRF \cite{Song2017TIP})
on fixed size regions. Furthermore, the multi-scale strategy is adopted
to capture more precise local information. However, these works either
encounter the problem of noise regions caused by predefined grids,
or simplify the spatial structural information, while our method can
explore the complex spatial structural layouts and reform them in
graphs to generate discriminative representations.

\subsection{Discriminative Region Discovery}

To discover the discriminative regions has been a long-standing study
in visual recognition. Singh \textit{et al.} \cite{DiscPat2012ECCV}
use an iterative optimization procedure to alternately clustering
and training discriminative classifier on densely sampled patches.
Juneja \textit{et al.} \cite{Juneja2013CVPR} first propose an initial
set of regions based on low-level segmentation cues, and then learn
detectors on top of these regions. However, these works all use the
handcrafted features as region representations.

Recently, Some works take advantage of CNN activations as region descriptors
for discriminative region discovery. Wu \textit{et al.} \cite{MetaObject2015ICCV}
obtain region proposals by performing MCG, and screen the regions
by using one-class SVM and RIM clustering. Cheng \textit{et al.}
\cite{Cheng2018PR} sample a set of local patches in a uniform grid
with their object scores extracted from ImageNet-CNN, then discard
the patches containing non-discriminative objects by applying Bayes
rules. One common characteristic of these works is that they generate
the candidate regions independently of the CNN classifiers, which
will incur much additional computational cost.

Besides these aforementioned approaches, some recent works explore
the convolutional responses from CNNs to directly discover discriminative
regions for fine-grained object recognition. Zheng \textit{et al.}
\cite{Zheng2017ICCV} group the convolutional channels to localize
object parts in the well constrained spatial configurations. Wei \textit{et al.}
\cite{Wei2017TIP} use a simple thresholding method to discover object
parts and select the largest component to represent the desired foreground
object. In contrast, we formulate the discovery procedure for scene
recognition, where more complex semantic regions and unconstrained
spatial structures exist. Similarly, the work of \cite{AdaDisc2018ACMMMM}
also uses a pretrained CNN classifier to generate discriminative regions
for scene images. However, it needs extra scene category cue for each
image and the CNNs with a specific architecture.

\subsection{Graph Neural Networks in Computer Vision}

Graph Neural Networks (GNNs) are designed to deal with the graph structured
data, which were first proposed in \cite{GNN2009TNN}. Recently, some
variants have been applied in program verification \cite{GGNN2016ICLR},
molecular property prediction \cite{MPNN2017ICML}, document classification
\cite{GCN2017ICLR} and made significant progress. Inspired by the
success of GNNs on graph structured data, some researches apply them
in computer vision task, like multi-label classification \cite{Marino2017CVPR},
situation recognition \cite{Li2017ICCV}, scene graph generation \cite{GraphRCNN2018ECCV},
zero-shot recognition \cite{Wang2018CVPR}, and etc. These works apply
GNNs to natural graph data (like knowledge graph \cite{Marino2017CVPR,Li2017ICCV,Wang2018CVPR}),
or constructed graph data with the supervision of annotated object
regions (like scene graph \cite{GraphRCNN2018ECCV}). In contrast
to them, we perform GCN \cite{GCN2017ICLR}, a variant of GNN, on
the structural layouts in scene images without external knowledge
or object annotations.

\section{Our Approach}

In this section, we first introduce how to construct Prototype-agnostic
Scene Layout (PaSL) from pretrained CNNs in an unsupervised way. Then
we build Layout Graph Network upon PaSL to integrate structural information
into visual representations. In the following, we will go into details
about our approach.

\subsection{Prototype-agnostic Scene Layout Construction}

PaSL is constructed by the locations and sizes of discriminative regions
(including objects, object-parts, and other visual patterns) in each
image. To form PaSL, we first need to discover discriminative regions.
Unlike previous works that use many selected image patches (from manual
annotation \cite{Izadinia2014CVPR} or region proposal \cite{MetaObject2015ICCV})
to train region detectors, we only need the convolutional units from
a pretrained CNN, without detector training.

Recently, Zhou \textit{et al.} \cite{Zhou2015ICLR} have shown the
convolutional units from a CNN pretrained on Places \cite{Places205NIPS2014}
dataset can be used as object detectors. And Bau \textit{et al.}
\cite{NetDissec2017CVPR} extend this conclusion to more pretrained
CNNs and more visual concepts. They demonstrated the individual convolutional
units in CNN can be aligned with semantic concepts across a range
of objects, parts, textures, scenes, materials, and colors. Inspired
by these works, we utilize the convolutional units in pretrained CNNs
as region detectors. In practice, given an image, we feed it into
a pretrained CNN to extract the convolutional activation maps $\mathcal{A}$
($\mathcal{A}\in R^{H\times W\times C}$) from the last convolutional
layer (For VGG16, max pooling need to be employed). The $c$-th activation
map in $\mathcal{A}$ is represented as $\mathcal{A}_{c}\in\mathbb{R}^{H\times W}$,
while $c\in\{1,\cdots,C\}$. For instance, if the resolution of the
input image is $224\times224$, we obtain $7\times7\times512$ activation
maps as $\mathcal{A}$, where $H=W=7$ and $C=512$, by adopting a
pretrained VGG16 model.

Based on the same assumption of \cite{Zhou2015ICLR,NetDissec2017CVPR}
that the desired regions (e.g., semantic regions) in feature maps
have high response values, we propose an adaptive threshold $T$ in
Eq.\ref{eq:thre_IbSL} to detect the candidates of discriminative
regions.

\begin{align}
T=\frac{1}{C}\sum_{c=1}^{C}\dot{\mathcal{A}}_{c},\ \ \dot{\mathcal{A}}_{c} & =\max\left(\mathcal{A}_{c}\right)\label{eq:thre_IbSL}
\end{align}
For efficient computing, any activation map whose maximum value is
under $T$ is discarded, then a subset $\widetilde{A}$ of activation
maps $\mathcal{A}$ is produced. Each activation map in $\widetilde{A}$
is scaled up to the input image resolution and then thresholded into
a binary map $B$ by using the threshold $T$. We take the connected
components in $B$ as the candidates of discriminative regions. The
algorithm from \cite{suzuki1985topological} is adopted to generate
bounding boxes of the connected components in each binary map. By
performing the same operations on all activation maps in $\widetilde{A}$,
we obtain bounding box set $M$ of all candidates of discriminative
regions. The element $m$ in $M$ is composed of the left-up and right-bottom
coordinates, e.g., $m=\{x_{min},\thinspace y_{min},\thinspace x_{max},\thinspace y_{max}\}$,
where $(x_{min},\thinspace y_{min})$ denotes the coordinate of left-up
point in bounding box, and $(x_{max},\thinspace y_{max})$ is the
coordinate of right-bottom point.

\begin{figure*}
\noindent \begin{centering}
\includegraphics[width=1\textwidth]{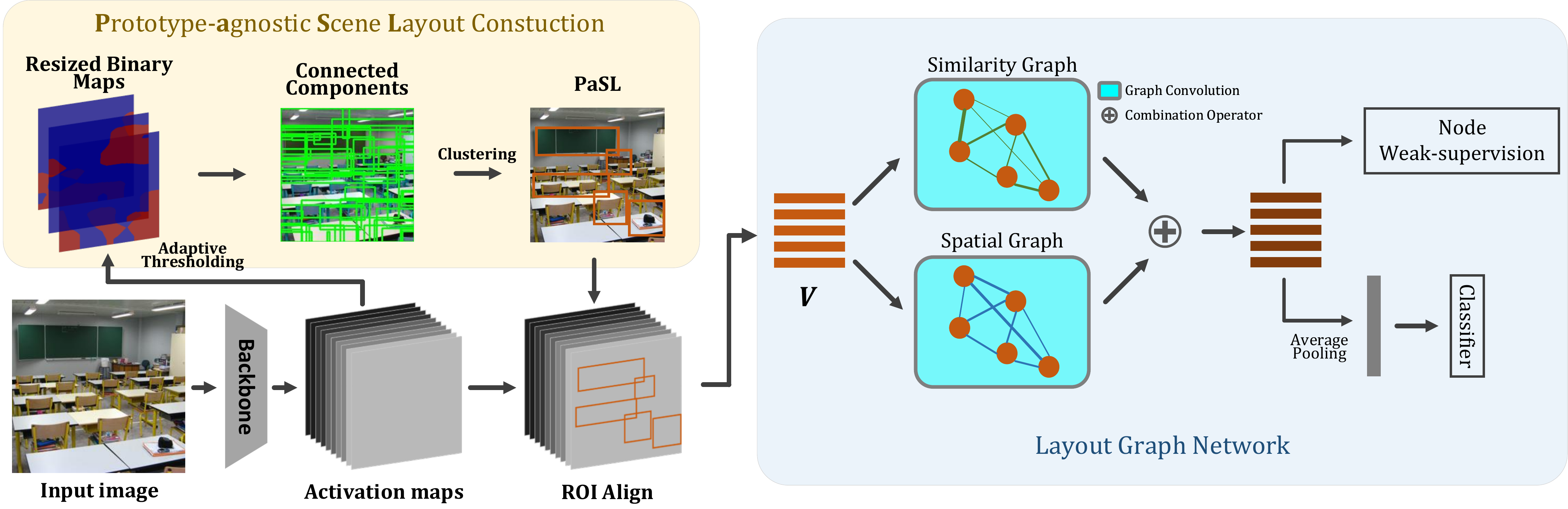}
\par\end{centering}
\noindent \centering{}\caption{\label{fig:The-structure-of-SLGN}Overview of our approach. A pretrained
CNN model (like VGG16 or ResNet50) is used as the backbone to extract
the candidates of discriminative regions. The desired discriminative
regions are clustered from these candidates, and fed into ROI Align
Layer to generate the node representations. Two subgraphs are constructed
by treating regions as nodes and designing spatial or similarity relation
as edge. Then, we perform graph convolution on two subgraphs and combine
them to obtain the final node representations. Finally, node weak-supervision
mechanism makes each node predict global image category by feeding
it into a fully-connected layer. Meanwhile, the averagely pooled node
representations, regarded as the global image representations, are
exploited for scene recognition by a fully-connected layer. }
\end{figure*}

In practice, the number of elements in $M$ is large, e.g., $\sim500$
for VGG16 and $\sim1500$ for ResNet50. If we construct PaSL with
all regions from $M$, it will cause expensive computational cost
in the later process. Meanwhile, the regions from $M$ have two characteristics.
One is that although adaptive thresholding can discard some small
noise parts, there also have several wrong detected results imposed
by the unsupervised process. Another one is that discovering from
each activation map independently may bring many visually similar
regions. In order to avoid the wrong or similar regions, we choose
a simple yet effective way, e.g., clustering, to find the most representative
regions in $M$ as the desired discriminative regions. Accordingly,
the discriminative regions $D$ could be obtained by:

\begin{align}
L & =\mathbb{C}(M,\thinspace N)\\
D & =\mathit{Aggre}(M,\thinspace L)
\end{align}
where $\mathbb{C}(\cdot)$ denotes hierarchical clustering method.
$N$ stands for the number of clusters, which also means the number
of discriminative regions. $L$ corresponds to the cluster labels
of the elements in $M$. Given cluster labels $L$, we perform mean
pooling method ($\mathit{Aggre}$) on bounding boxes of elements in
the same cluster to obtain bounding boxes of discriminative regions
$D=\{d_{1},\thinspace\dots,\thinspace d_{N}\}$. Specifically, clustering
method can be k-means or spectral clustering. If choosing k-means,
cluster centers can be directly used as discriminative regions.

Given discriminative regions, we define Prototype-agnostic Scene Layout
(PaSL) as a collection of the locations and sizes of these regions
in each image. The whole process is shown in Fig.\ \ref{fig:The-structure-of-SLGN}.
The spatial structure, that is implicit in PaSL, requires to be represented
in a certain form. To form the diverse and free topological structure
of PaSL of each image, the graph is adopted as data structure. Following
the common setting of graph structured data, we define the discriminative
regions as nodes and encode two kinds of independent relations between
regions as edges. The details will be described in the following section.

\subsection{Layout Graph Network}

For modeling the spatial structure of PaSL, we reorganize it as a
layout graph, which is better for incorporating the structural information
into visual representations. Given PaSL with discriminative regions,
a layout graph $\mathcal{G}=\{\mathit{V},\thinspace\mathbf{\mathit{A^{sp}}},\thinspace\mathit{A^{sim}}\}$
is constructed, which contains a node set $\mathit{V}$, and two adjacency
matrices $\mathit{A^{sp}}$, $\mathit{A^{sim}}$. For clarity, we
decompose the layout graph into two subgraphs with the same nodes
but different adjacency matrices: spatial subgraph $\mathcal{G}^{sp}=\{\mathit{V},\thinspace\mathit{A}^{sp}\}$
and similarity subgraph $\mathcal{G}^{sim}=\{\mathit{V},\thinspace\mathit{A}^{sim}\}$.
More specifically, these two subgraphs share the same node set $V=\{v_{1},\thinspace\dots,\thinspace v_{N}\}$,
where $v_{i}$ corresponds to the representation of discriminative
region $i$. We apply RoIAlign \cite{MaskRCNN2017ICCV} to extract
the representation of each region from a pretrained CNN as the initial
state vector of $v_{i}$. This pretrained CNN can be regarded as a
feature extraction model, which is same as the pretrained model for
generating PaSL, unless otherwise stated.

\textbf{Spatial Subgraph.} The spatial information is vital in PaSL,
because it implies the functions or properties of regions. One way
to take advantage of this information is to exploit the spatial relations
between regions. Specifically, we define a kind of spatial representation
to encode this relation and then generate the spatial edge to form
the adjacency matrix. As mentioned above, each discriminative region
$i$ has a bounding box $d_{i}=\{x_{min},\thinspace y_{min},\thinspace x_{max},\thinspace y_{max}\}$.
Inspired by \cite{Yu2017ICCV}, we extract the spatial feature of
each region as follows:

\begin{equation}
d_{i}^{sp}=\left[\frac{x_{min}}{W_{img}},\thinspace\frac{y_{min}}{H_{img}},\thinspace\frac{x_{max}}{W_{img}},\thinspace\frac{y_{max}}{H_{img}},\thinspace\frac{Ar}{Ar_{img}}\right]
\end{equation}
where $Ar$ and $Ar_{img}$ are the areas of the region $i$ and the
image respectively. $W_{img}$ and $H_{img}$ denotes the width and
height of the image. We concatenate $d_{i}^{sp}$ and $d_{j}^{sp}$
to obtain the spatial representation $d_{ij}^{sp}$ of spatial relation
between the regions $i$ and $j$. After generating the spatial representation,
we employ an edge function $F_{e}$, implemented as one-layer fully
connected network, to generate the spatial edge $a_{ij}^{sp}$ as:

\begin{align}
e_{ij}^{sp} & =F_{e}(d_{ij}^{sp})\\
a_{ij}^{sp} & =\frac{\exp(e_{ij}^{sp})}{\sum_{j\neq i}\exp(e_{ij}^{sp})}
\end{align}
Then the spatial adjacency matrix $A^{sp}$ is obtained to form spatial
subgraph $\mathcal{G}^{sp}$. The diagonal values in $A^{sp}$ are
zero.

\textbf{Similarity Subgraph.} To explore the spatial information
in PaSL is an obvious requirement. But there exists an problem in
the spatial subgraph that the spatial relation overlooks the semantic
meanings of regions. To address this problem, we propose the similarity
subgraph as a complement to the spatial subgraph. Due to the lack
of explicit labels for the local regions, we take the region representation
as a substitution for the semantic label. Then, we model the similarity
between these region representations to capture the semantic similarity
relations between regions.

Given the node set $V$, we can obtain the state vector $v_{i}\in\mathbb{R}^{h}$
of each node. In similarity subgraph, we aim to obtain the strong
connection between semantic similar regions. So the semantic similarity
relations between regions are measured by the cosine similarity, which
is defined as follows:

\begin{equation}
e_{ij}^{sim}=\phi(v_{i})^{T}\phi(v_{j}),\ \ \phi(v)=\frac{\omega^{sim}v}{\left|\left|\omega^{sim}v\right|\right|_{2}}
\end{equation}
where $\phi$ represents the transformation of the state vector and
following $\ell_{2}$ normalization, $\omega^{sim}\in\mathbb{R}^{h/2\times h}$
is the transformation weights. The dot product $e_{ij}^{sim}$ of
two $\ell_{2}$ normalized vector denotes the cosine similarity between
regions. To balance the impact of neighbor nodes, we perform the softmax
function on each row of the cosine similarity matrix as:

\begin{equation}
a_{ij}^{sim}=\frac{\exp(e_{ij}^{sim})}{\sum_{j\neq i}\exp(e_{ij}^{sim})}
\end{equation}
where $A^{sim}$ is used as the adjacency matrix for similarity subgraph.
The diagonal values in $A^{sim}$ are zero.

\textbf{Graph Convolution.} After building the layout graph, the next
step is to incorporate the spatial and semantic similarity information
into the representations of regions, and generate discriminative image
representations. Considering the superior performance of Graph Convolution
Network (GCN) \cite{GCN2017ICLR} on graph structured data, we adopt
Graph Convolution (GC) on spatial and similarity subgraph, then combine
two subgraphs. Given a graph $\mathcal{G}=\{V,\thinspace A\}$, where
$V\in\mathbb{R}^{N\times h}$ is the node set and $A\in\mathbb{R}^{N\times N}$
is the adjacency matrix. One GC layer aims to combine the information
of neighbor nodes and target node through relation edges to update
the state vector of target node, which can be formulated as:

\begin{align}
V^{t} & =\sigma(ZV^{t-1}\Omega_{t})
\end{align}
\[
Z=\Lambda^{-1}\tilde{A}
\]
\[
\Lambda_{ii}=\sum_{j}\tilde{A}_{ij},\ \ \tilde{A}=A+I_{N}
\]
where $V^{t}$ is the updated state vectors of nodes in GC layer $t$,
$\Omega_{t}\in\mathbb{R}^{h_{t-1}\times h_{t}}$ denotes weight matrix,
and $h_{0}$ is the input vector dimension while $h_{t}\thinspace(t>0)$
means hidden size of GC layer $t$. We utilize the non-linear function
ReLU as $\sigma$.

\textbf{Combination of Different Subgraphs.} Now, we can employ graph
convolution to generate the updated state vectors $V_{sim}^{t}$,
$V_{sp}^{t}$ for spatial, similarity subgraphs, with $A^{sim}$,
$A^{sp}$ obtained above, respectively. Then, we investigate how to
effectively combine these two subgraphs. First, we define the combination
of two subgraphs as:

\begin{equation}
V^{t}=V_{sim}^{t}\oplus V_{sp}^{t}
\end{equation}
where $\oplus$ means the combination operator. Intuitively, we can
combine the updated state vectors from two subgraphs by using element-wise
\textit{addition} or \textit{maximum}. Beyond them, we also consider
an alternative to improve the sparsity of combined representations,
which is element-wise \textit{product}. we conduct a comparison experiment
in section \ref{subsec:Effect-of-different}, which confirms that
the element-wise \textit{product} is a better choice to combine two
subgraphs.

\textbf{Global information.} PaSLs in most images cannot cover the
whole areas of images, which may lose some useful information. So
we decide to add global information into the layout graph. We define
a global node that represents the whole image, and perform average
pooling on the convolutional activation maps from the last convolutional
layer to generate the initial state vector of the global node. As
a result, the node set $V$ will be $\{v_{0},v_{1},\cdots,v_{N}\}$,
where $v_{0}$ denotes the global node. For spatial subgraph, We set
the bounding box of global node as $d_{0}=\{x_{min}=0,\thinspace y_{min}=0,\thinspace x_{max}=W_{img},\thinspace y_{max}=H_{img}\}$,
where $W_{img}$ and $H_{img}$ denote the width and height of the
whole image. The global node is connected to all local nodes, and
we apply the same operations described above to obtain the new adjacency
matrices $A^{sp}$, $A^{sim}$.

\textbf{Output.} To avoid overfitting, we only utilize one GC layer.
We obtain the final state vectors $V^{1}\in\mathbb{R}^{N\times h_{1}}$
from the GC layer and following $\ell_{2}$ normalization as node
representations. When only using local regions as nodes, we apply
average pooling on node representations to generate the image representation
as a $h_{1}$-dimensions vector. And if adding global node, we only
treat the global node representation as the image representation.
Besides, we have tried to averagely pool all global/local node representations
to obtain the image representation, which hurts the performance. And
we have also tried to concatenate the global node representation with
averagely pooled local node representation to produce the image representation,
while it has similar performance but needs more parameters in the
later process. For scene recognition, we feed image representation
into one layer fully connected network to predict the image category.
And we utilize softmax function with cross entropy as the loss function
to obtain the image classification loss $l_{g}$.

\textbf{Node Weak-supervision Mechanism.} Specifically, we propose
a node weak-supervision mechanism to improve the discriminative performance
of each node (except global node). For the representation of each
node, we force it to predict the scene category of image by using
one layer fully-connected network in a weakly supervised way, which
can make the node representations more suitable for image recognition
and produce the node classification loss $l_{n}$. We combine the
two classification loss to form the total loss $l$ as,

\begin{equation}
l=l_{g}+\lambda l_{n}
\end{equation}
where $\lambda$ is a hyperparameter. Specifically, this branch is
only used in the training process.

\section{Experiments and Discussions}

In this section, we evaluate our method on three widely recognized
datasets, MIT67 \cite{MIT67CVPR2009}, SUN397 \cite{SUN397CVPR2010},
and Places365 \cite{Places2TPAMI2018}.

\textbf{MIT67 Dataset} contains a total of 15620 images belonging
to 67 indoor scene categories. Following the standard evaluation protocol,
we use 80 images of each category for training and 20 images for testing.
We report accuracy as evaluation metric.

\textbf{SUN397 Dataset} is a more challenge scene dataset, which contains
397 scene categories and 108,754 images. The dataset is divided into
10 train/test splits, each split consists of 50 training images and
50 test images per category. The average accuracy over splits is presented
as evaluation metric.

\textbf{Places365 Dataset} is one of the largest scale scene-centric
datasets, which has two training subsets, Places365-standard and Places365-challenge.
In this paper, we only choose Places365-standard as training set,
which consists of around 1.8 million training images and 365 scene
categories. The validation set of Places365 contains 100 images per
category and the testing set has 900 images per category. We report
experimental results on its validation set, because its test set has
no available ground truth. Both top1 and top5 accuracy are reported
as evaluation metric.

\subsection{Implementation Details}

Our model can be implemented with different pretrained models as backbone
CNNs. For fair comparison with other methods, we adopt three pretrained
models, which are VGG-IN, VGG-PL205, ResNet-PL365. VGG-IN, VGG-PL205
are the VGG16 models pretrained on ImageNet dataset \cite{ImageNet2009CVPR}
and Places205 dataset \cite{Places205NIPS2014} respectively, ResNet-P365
is the ResNet50 model pretrained on Places365 dataset \cite{Places2TPAMI2018}.
To construct PaSL, we extract the convolutional activation maps from
the last convolutional layer (max-pooled in VGG16). Inspired by \cite{Herranz2016CVPR},
we fix the input image resolution as $448\times448$ for VGG16-IN
and ResNet50-PL365, $352\times352$ for VGG16-PL205, which leads to
$14\times14\times512$, $14\times14\times2048$ and $11\times11\times512$
activation maps respectively. The number of clusters $N$, the hidden
size $h_{1}$ and the $\lambda$ are set to $\{32,8192,4.0\}$ for
LGN with backbone VGG-IN and VGG-PL205, $\{64,\thinspace8192,\thinspace1.0\}$
for ResNet-PL365.

The initial state vectors of nodes are normalized with two normalization
function (Layer Normalization \cite{LayerNorm2016arXiv}, $\ell_{2}$
Normalization), then fed into LGN. Specifically, the Layer Normalization
is not trained in our experiments. We train LGN using Adam \cite{kingma2014adam}
with an initial learning rate of $10^{-3}$ (decayed by a factor of
0.1 at 10/15/18th epoch), a batch size of 32 and weight decay of $10^{-5}$.
All parameters are randomly initialized following Xavier initialization
method \cite{GlorotInitializer2010}. We use the model trained at
20th epoch as the final model in all experiments. Dropout is only
applied on the output prediction layer with a ratio of $0.2$. The
$\ell_{2}$-norm of gradients is clipped to a maximum value of $0.25$.
All experiments are conducted on a single NVIDIA 1080 Ti GPU by using
open-sourced framework Tensorflow.

\subsection{Experimental Results}

In this subsection, we first report the performances on MIT67, SUN397.
These two datasets are the most popular benchmark for evaluating scene
recognition methods. Thus, we can provide the comprehensive and detailed
comparison with existing works about scene recognition. Meanwhile,
we also conduct experiments on one of the largest scale scene dataset,
Places365, to demonstrate the generalization of our model.

\subsubsection{Comparison on single model in single scale (MIT67 and SUN397)}

Most existing scene recognition methods obtain their best performances
based on multi-model or multi-scale fusion. However, to perform the
fusion needs more computational time and memory usage, which will
cause expensive cost. The idea of multi-scale representation is presented
to alleviate the problem of the various sizes of the semantic components
in scene images. Benefiting from the flexible structure of PaSL, our
model can efficiently capture the different locations and sizes of
semantic components, to produce the better image representations for
scene recognition. To prove it, we compare the previous works with
our model on single VGG16 model in single scale in Table \ref{tab:Comparison-of-Single-model}.
The two pretrained VGG16 models, pretrained on ImageNet (VGG-IN) and
Places205 (VGG-PL205) are adopted as backbone CNN models in the comparison.
The backbone VGG-PL205 show impressive performance on MIT67 and SUN397,
generally outperforming the VGG-IN. Compared to existing works using
the same VGG-PL205 backbone, our model obtains better performance
with a clear margin ($1-2\%$). While based on the VGG-IN, the LGN
surpasses the most previous works, except MFAFVNet and LSO-VLADNet.
The lower performance of VGG-IN can be concluded into two possible
reasons: 1) these two previous works report better accuracy benefiting
from the refinement of low level convolutional features. 2) The PaSL
derived from VGG-IN has less power for capturing the spatial structure
in scene images, which is verified in subsection \ref{The-Generalization-of-PaSL}.

\begin{table}
\noindent \begin{centering}
\caption{\label{tab:Comparison-of-Single-model}Comparison of LGN with previous
methods based on single VGG16 model in single scale. Classification
accuracy (\%) is reported as evaluation metric on MIT67 and SUN397
datasets. The best result of each column is marked in bold.}
\par\end{centering}
\noindent \centering{}%
\begin{tabular}{ccccc}
\hline
Method & P.S. & I.R. & MIT67 & SUN397\tabularnewline
\hline
VGG16 \cite{Herranz2016CVPR} & IN & 643 & 76.42 & 59.71\tabularnewline
MFA-FS \cite{Dixit2016NIPS} & IN & 512 & 79.57 & 61.71\tabularnewline
HSCFVC \cite{SCFVC2017TPAMI} & IN & 512 & 79.5 & -\tabularnewline
MFAFVNet \cite{MFAFVNet2017ICCV} & IN & 512 & 80.3 & 62.51\tabularnewline
CNN-DL \cite{Liu2018AAAI} & IN & - & 78.33 & 60.23\tabularnewline
LSO-VLADNet \cite{Chen2018PR} & IN & 448 & 81.7 & 61.6\tabularnewline
VGG16 \cite{Herranz2016CVPR} & PL205 & 256 & 80.90 & 66.23\tabularnewline
S-HunA \cite{Sicre2017CVPR} & PL205 & - & 83.7 & -\tabularnewline
SpecNet \cite{Khan2017ICCV} & PL205 & 256 & 84.3 & 67.6\tabularnewline
CNN-DL \cite{Liu2018AAAI} & PL205 & - & 82.86 & 67.90\tabularnewline
\hline
\multirow{2}{*}{LGN} & IN & 448 & 79.78 & 62.03\tabularnewline
 & PL205 & 352 & \textbf{85.37} & \textbf{69.48}\tabularnewline
\hline
\multicolumn{5}{c}{P.S. : ``Pretrain dataset'', I.S. : ``Input Resolution''}\tabularnewline
\multicolumn{5}{c}{IN : ``ImageNet'', PL205 : ``Places205''}\tabularnewline
\end{tabular}
\end{table}

\subsubsection{Comparison with the state-of-the-art works (MIT67 and SUN397)}

Table \ref{tab:Comparison-of-SoTA} presents the results of our best
model and state-of-the-art works. Our best model is based on ResNet-PL365
pretrained model in single scale setting. Compared to the methods
\cite{AdaDisc2018ACMMMM,PowerNorm2018CVPR,Dixit2019TPAMI} based on
the same pretrained model, our model achieves the best performance.
Most importantly, the work \cite{AdaDisc2018ACMMMM} utilizes the
similar technique to extract discriminative regions and even multi-scale
regions to generate the image representations. However, it ignores
the relations (either spatial or similarity relations) between local
regions, leading to an inferior performance. This confirms that the
relations between local regions are useful for scene recognition,
and our LGN can take advantage of them. We also report state-of-the-art
works that involve various combination techniques to achieve better
performance. Even though these works contain multi-scale information
\cite{CLDL2016ECCV,Dixit2016NIPS,MFAFVNet2017ICCV,Liu2018AAAI,Cheng2018PR,AdaDisc2018ACMMMM,Pan2019TIP}
or multi-model combination \cite{Dixit2016NIPS,MFAFVNet2017ICCV,Cheng2018PR,Dixit2019TPAMI},
our model still outperforms them and achieves the state-of-the-art
performance for scene recognition, to the best of our knowledge.

\begin{table}
\noindent \begin{centering}
\caption{\label{tab:Comparison-of-SoTA}Comparison of LGN with state-of-the-art
works on MIT67 and SUN397. Classification accuracy (\%) is reported
as evaluation metric. The best result of each column is marked in
bold.}
\par\end{centering}
\noindent \centering{}%
\begin{tabular}{ccc}
\hline
Method & MIT67 & SUN397\tabularnewline
\hline
CLDL \cite{CLDL2016ECCV} & 84.69 & 70.40\tabularnewline
MFA-FS \cite{Dixit2016NIPS} & 87.23 & 71.06\tabularnewline
MFAFVNet \cite{MFAFVNet2017ICCV} & 87.97 & 72.01\tabularnewline
CNN-DL \cite{Liu2018AAAI} & 86.43 & 70.13\tabularnewline
SDO \cite{Cheng2018PR} & 86.76 & 73.41\tabularnewline
PowerNorm \cite{PowerNorm2018CVPR} & 86.3 & -\tabularnewline
fgFV \cite{Pan2019TIP} & 87.60 & -\tabularnewline
MFAFSNet \cite{Dixit2019TPAMI} & \textbf{88.06} & 73.35\tabularnewline
Adi-Red \cite{AdaDisc2018ACMMMM} & - & 73.59\tabularnewline
\hline
LGN (ResNet) & \textbf{88.06} & \textbf{74.06}\tabularnewline
\hline
\end{tabular}
\end{table}

\subsubsection{Experimental results on Places365}

To make more convincing results, we report the result of our best
model on Places365 in Table \ref{tab:Classification-Places365}. The
experimental setting is same as above, except the input resolution
changes to $224\times224$ and the number of clusters changes to $32$.
Compared to the baseline Places365-ResNet \cite{Places2TPAMI2018},
our model can gain 1.76\% improvement of Top1 accuracy, which demonstrates
the effectiveness of the proposed PaSL and LGN. It is worthy to note
that the proposed LGN can outperform previous works with single model
in single scale, although they report better results obtained by multi-model
or multi-scale combination.

\begin{table}
\caption{\label{tab:Classification-Places365}Classification accuracy (\%)
on Places365 validation set.}
\noindent \centering{}%
\begin{tabular}{c|c|c}
\hline
\multicolumn{1}{c}{Method} & \multicolumn{1}{c}{Top1 acc.} & Top5 acc.\tabularnewline
\hline
Places365-VGG \cite{Places2TPAMI2018} & 55.24 & 84.91\tabularnewline
Places365-ResNet \cite{Places2TPAMI2018} & 54.74 & 85.08\tabularnewline
Deeper BN-Inception \cite{Wang2017TIP} & 56.00 & 86.00\tabularnewline
CNN-SMN \cite{Song2017TIP} & 54.3 & -\tabularnewline
LGN (ResNet) & 56.50 & 86.24\tabularnewline
\hline
Multi-Model CNN-SMN \cite{Song2017TIP} & 57.1 & -\tabularnewline
Multi-Resolution CNNs \cite{Wang2017TIP} & 58.30 & 87.60\tabularnewline
\hline
\end{tabular}
\end{table}

\subsection{Analysis of PaSL}

We provide a deep analysis of PaSL based on MIT67, and discuss its
properties.

\subsubsection{The Visualization of PaSL}

\begin{figure*}
\noindent \begin{centering}
\includegraphics[width=1\textwidth]{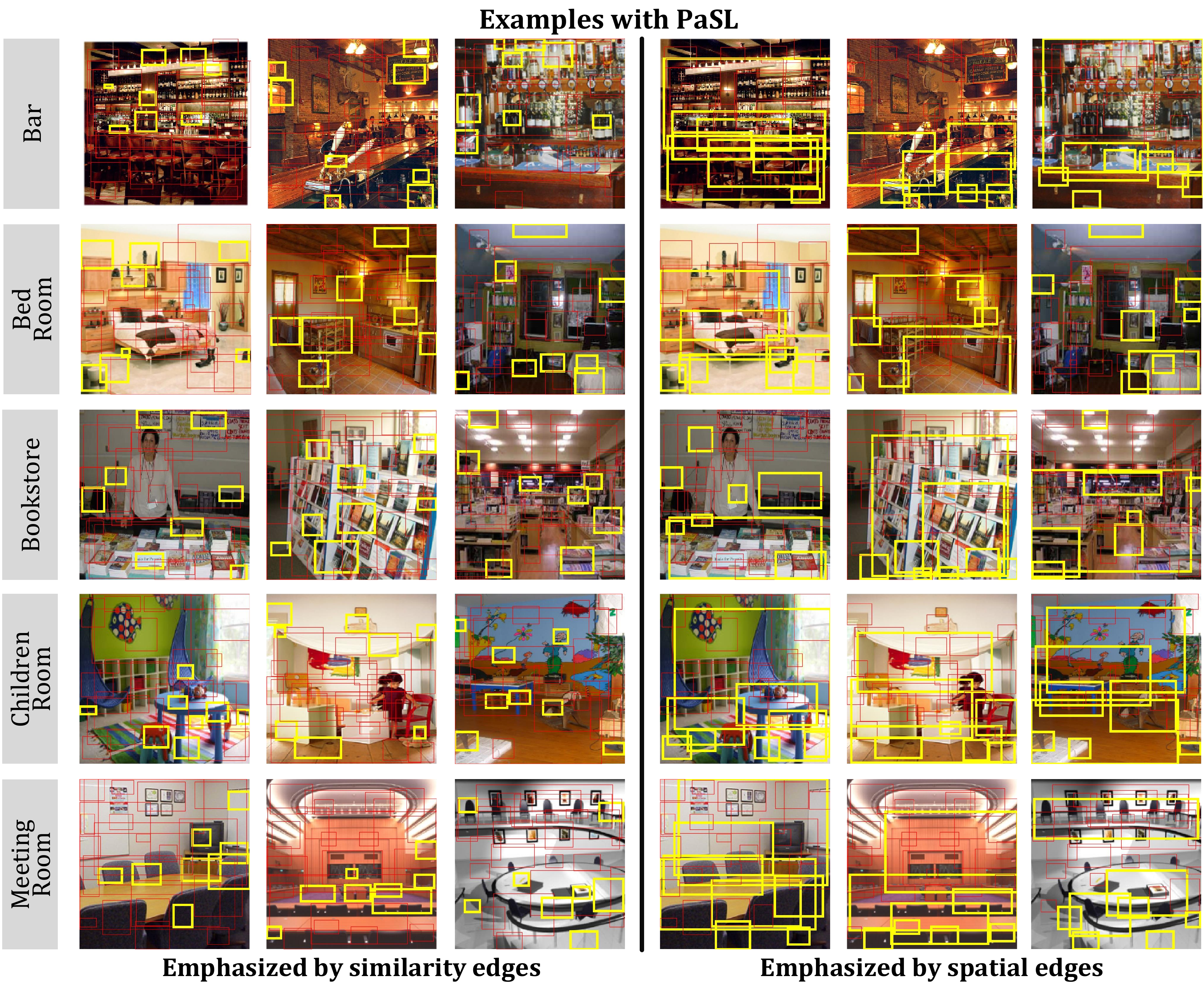}
\par\end{centering}
\caption{\label{fig:Visualization-of-PaSL}Visualization of PaSL. We choose
some examples with PaSL from five scene categories. Each example shows
the image with 32 rectangles representing the local regions in PaSL.
For better visualization, we sort all 32 local regions and emphasize
the top 8 regions in yellow and thick rectangles and downplay other
regions in red and thin rectangles. The left 3 columns show the images
with the regions emphasized by similarity edges. The right 3 columns
show the same images with the regions emphasized by spatial edges.
This figure is better viewed in color.}
\end{figure*}

Fig.\ \ref{fig:Visualization-of-PaSL} show the images with PaSL
derived from the backbone VGG-PL205. All the images are plotted with
32 bounding boxes of regions in PaSL. To avoid an unclear display,
we firstly sort all the local regions in PaSL, and then emphasize
the top 8 regions in the yellow and thick rectangles and downplay
other regions in the red and thin rectangles, when plotting PaSL on
an image. Specifically, we choose the edge values of all local regions
connected to the global node in two adjacency matrices for sorting
these regions. In Fig.\ \ref{fig:Visualization-of-PaSL}, the left
3 columns show the regions emphasized by similarity edges, and the
right 3 columns show the regions emphasized by spatial edges in same
images. It's easy to see that the regions in PaSL can vary greatly
in size and location to suit the large diversity of structural layouts
in scene images. Importantly, PaSL can localize some semantic regions
specified for the corresponding scenes, like ``liquor cabinet''
in ``bar'', ``bed'' in ``bed room'', ``meeting table'' in
``meeting room'', and so on. When comparing the regions emphasized
by spatial and similarity edges, the obvious difference is that the
regions emphasized by spatial edges tend to focus on the aggregated
semantic components (like a lots of chairs), and the regions emphasized
by similarity edges usually concentrate on the contents similar in
visual details (like texture of parts of floor or wall). This difference
demonstrates that two subgraphs can explore the local information
in different aspects and be complementary to each other.

\subsubsection{The Difference of PaSL}

\begin{figure}
\begin{centering}
\includegraphics[width=0.7\columnwidth]{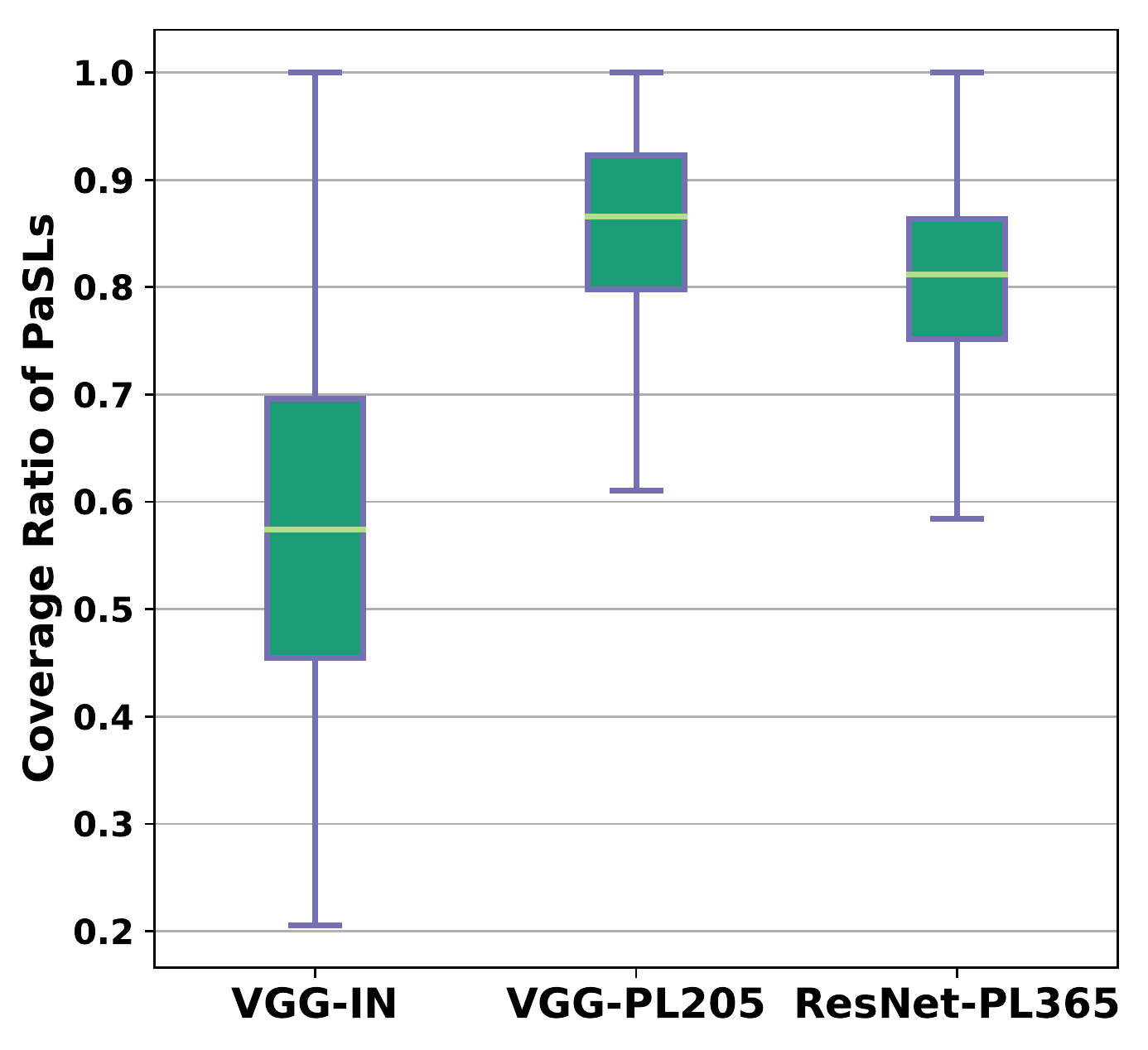}
\par\end{centering}
\caption{\label{fig:The-cover-ratio}The distributions of Coverage Ratio of
PaSLs derived from different pretrained models (VGG-IN, VGG-PL205,
and ResNet-PL365). Coverage Ratio is the ratio between the coverage
area of PaSL and the area of the image.}
\end{figure}

Although each image has its own spatial structure, PaSLs derived from
the same pretrained model will have some similar properties. From
the point of view of PaSLs in the whole training data, we define a
metric named Coverage Ratio, which is the ratio between the coverage
area of PaSL and the area of the image, to analyze the properties
of PaSLs. In Fig.\ \ref{fig:The-cover-ratio}, the boxplots show
the distributions of Coverage Ratio for all training image PaSLs derived
from three different pretrained models. Note that the number of regions
in PaSL is fixed to 32 for a fair comparison. We find that PaSLs derived
from models pretrained on scene-centric datasets (Places205 or Places365)
focus on larger regions compared to them derived from the model pretrained
on object-centric dataset (ImageNet). And also PaSLs derived from
the model pretrained on ImageNet may focus on the regions with high
objectness. So, the values of their Coverage Ratio have a larger diversity
due to the wide variety of size and location of objects in scene images.

\subsubsection{The Generalization of PaSL}

\label{The-Generalization-of-PaSL}Considering the independence between
PaSL construction and LGN, we can explore the generalization of PaSL
by combining PaSL with LGN when they are based on different or same
pretrained models. There are three kinds of PaSLs derived from different
pretrained models, and three kinds of LGNs based on different pretrained
models. Therefore, we conduct combination experiments on MIT67, and
report nine combination results in Table \ref{tab:Generalization-of-SL}.
When the pretrained models are different for PaSL and LGN, the performance
can yield a change of no more than $1.04\%$. Besides different combination
of PaSL and LGN, we also evaluate another spatial layout formed by
regions generated by Faster RCNN \cite{FasterRCNN2017TPAMI} pretrained
on MSCOCO dataset. For a fair comparison, we set the number of regions
in this layout to 32. Based on Table \ref{tab:Generalization-of-SL},
we can have three observations. 1) Compared to PaSLs derived from
other pretrained models, the one from VGG-PL205 has the better ability
to represent the spatial structure of scene images. 2) Despite having
some fluctuations in performance, PaSLs derived from different pretrained
models have comparable value for scene recognition, which demonstrates
their considerable generalization capability. 3) The spatial layout
generated by object detection obtains the worst performances with
all LGNs. One possible reason is that this layout mainly focus on
some common objects, and is not suitable to capture the complex structural
layouts of scene images.

\begin{table}
\caption{\label{tab:Generalization-of-SL}The generalization of PaSLs derived
from different pretrained models. We show classification accuracy
(\%) of the combinations of PaSL and LGN, when they are based on different
or same pretrained models. The best result of each row is marked in
bold.}
\noindent \centering{}%
\begin{tabular}{c|cccc}
\hline
\multirow{2}{*}{\textbf{LGN}} & \multicolumn{3}{c|}{\textbf{PaSL}} & \multirow{2}{*}{Detection}\tabularnewline
\cline{2-4}
 & VGG-IN & VGG-PL205 & \multicolumn{1}{c|}{ResNet-PL365} & \tabularnewline
\hline
VGG-IN & 79.78 & \textbf{80.52} & 80.07 & 68.66\tabularnewline
VGG-PL205 & 84.78 & \textbf{85.37} & 85.29 & 80.52\tabularnewline
ResNet-PL365 & 88.21 & \textbf{88.73} & 87.69 & 83.66\tabularnewline
\hline
\end{tabular}
\end{table}

\subsection{Experimental Study of LGN}

\subsubsection{Configuration of Hyperparameters}

Three hyperparameters are important to determine the performance of
our method, the number of clusters $N$ in constructing PaSL, the
hidden size $h_{1}$ in graph convolution, and the $\lambda$ in node
weak-supervision mechanism. To investigate these three hyperparameters,
we conduct several experiments on MIT67 dataset. Because the architectures
of VGG16 and ResNet50 are different, especially the processes from
the last convolutional layer to output prediction layer, we analyze
these hyperparameters on VGG-PL205 and ResNet-PL365 pretrained models,
separately. We do not show the analysis on VGG-IN, since it has a
similar behavior with VGG-PL205.

\begin{figure}
\noindent \begin{centering}
\begin{tabular}{cc}
\includegraphics[width=0.45\columnwidth]{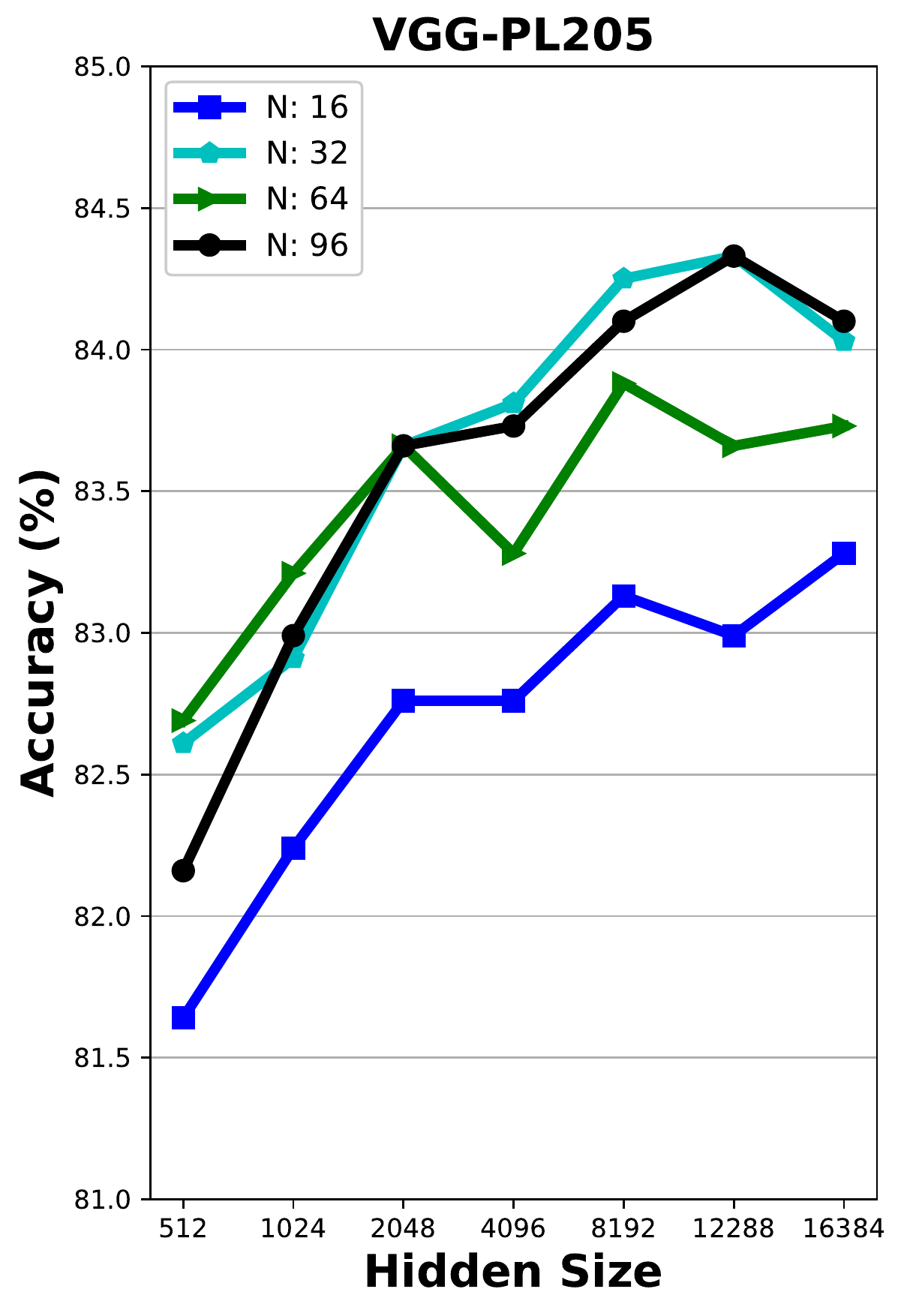} & \includegraphics[width=0.45\columnwidth]{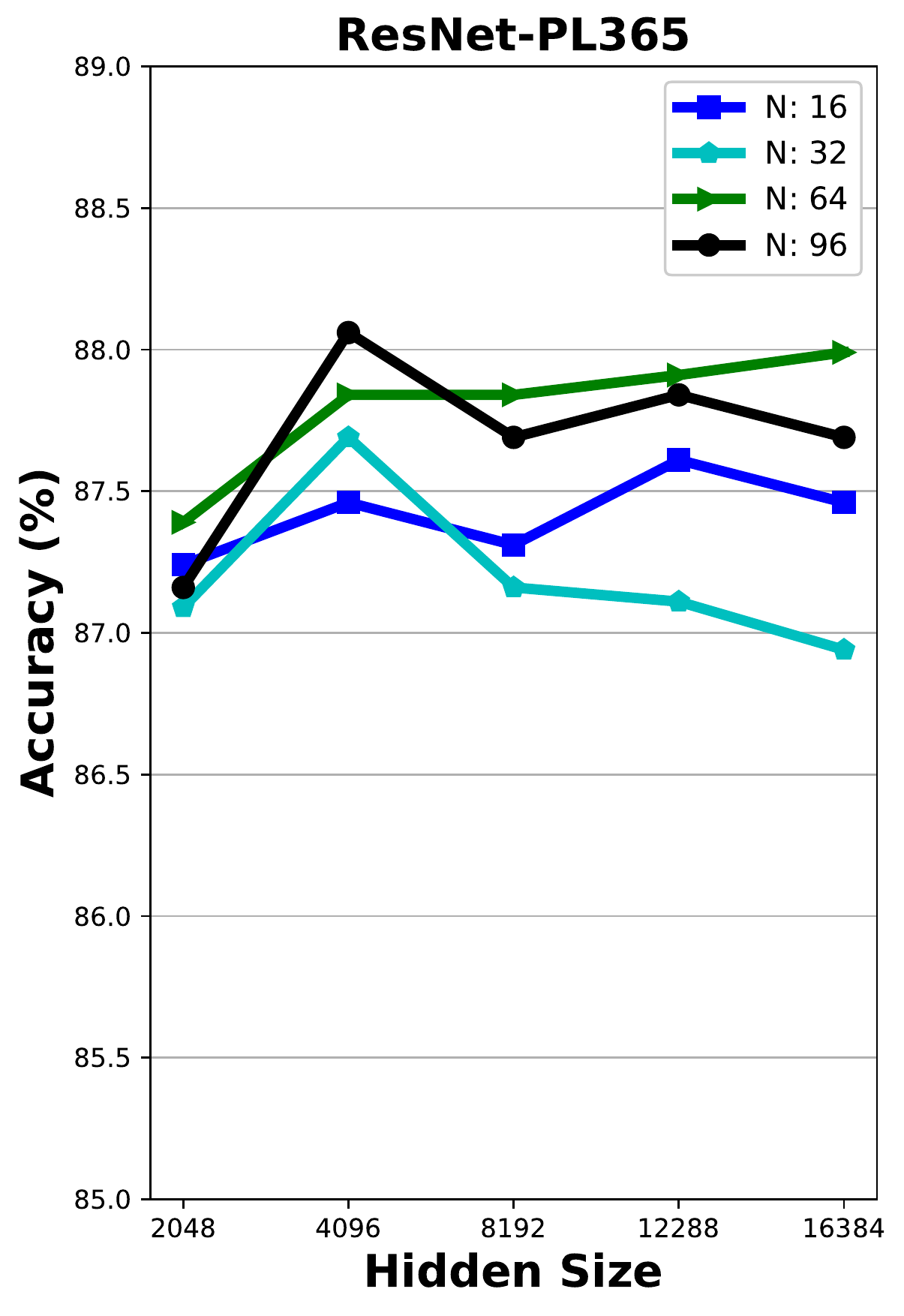}\tabularnewline
(a) & (b)\tabularnewline
\end{tabular}
\par\end{centering}
\noindent \centering{}\caption{\label{fig:The-effect-of-h-N}The effect of hidden size $h_{1}$ and
the number of clusters $N$. We report the results on spatial subgraph
without global information and node weak-supervision. Because of the
architecture difference of VGG16 and ResNet50, we show the analysis
results separately.}
\end{figure}

We evaluate the effect of hidden size $h_{1}$ and the number of clusters
$N$ on spatial subgraph without global information and node weak-supervision
in Fig.\ \ref{fig:The-effect-of-h-N}. It can be observed that the
trends of accuracy caused by hidden size $h_{1}$ are different with
VGG-PL205 and ResNet-PL365. In Fig.\ \ref{fig:The-effect-of-h-N}
(a), the accuracy has a significant increment when hidden size $h_{1}$
is lower than 8192, and then tends to be stable as hidden size $h_{1}$
increases. However, in Fig.\ \ref{fig:The-effect-of-h-N} (b), we
can see that the accuracy has a slight change as hidden size $h_{1}$
changes. These differences can be attributed to the aggregation techniques
for generating the global image representation in different CNNs.
In VGG16, the local spatial features are concatenated to produce the
global representations, while they are averagely pooled in ResNet.
Thus, in LGN based on VGG-PL205, aggregating the local features need
to substantially enlarge the projection dimension (hidden size $h_{1}$)
to prevent the information loss from averagely pooling, but not for
ResNet-PL365. For instance, the ratios of hidden size $h_{1}$ to
input dimension $h_{0}$ are 16 and 4 for VGG-PL205 and ResNet-PL365,
respectively.

As illustrated in Fig.\ \ref{fig:The-effect-of-h-N}, when the number
of clusters $N=32$, we obtain better performances by using VGG-PL205,
and the similar observation can be found at $N=64$ with ResNet-PL365.
Thus, we set hidden size and the number of clusters $\{h_{1},\thinspace N\}$
to $\{8192,\thinspace32\}$ and $\{8192,\thinspace64\}$ for VGG-PL205
and ResNet-PL365 respectively in all subsequent experiments. Besides
$h_{1}$ and $N$, the hyperparameter $\lambda$ in node weak-supervision
mechanism is also important. The node weak-supervision mechanism aims
to force each local node to predict the image category, which makes
local representations more specific for generating discriminative
image representations. We report the results on spatial subgraph without
global information for different values of $\lambda$ in Table \ref{tab:The-influence-of-lambda}.
It can be observed that, the best performances are obtained at $\lambda=4.0$
and $\lambda=1.0$ for VGG-PL205 and ResNet-PL365, respectively, which
are set as default hyperparameters in subsequent experiments. We set
the same hyperparameters $\{h_{1}=8192,\thinspace N=32,\thinspace\lambda=4.0\}$
for VGG-IN pretrained models.

\begin{table}
\noindent \begin{centering}
\caption{\label{tab:The-influence-of-lambda}The influence of $\lambda$ in
node weak-supervision mechanism. We show the classification accuracy
(\%) based on spatial subgraph without global information. The best
result of each row is marked in bold.}
\par\end{centering}
\noindent \begin{centering}
\begin{tabular*}{1\columnwidth}{@{\extracolsep{\fill}}ccccccc}
\hline
$\lambda$ & 0.0 & 1.0 & 2.0 & 3.0 & 4.0 & 5.0\tabularnewline
\hline
VGG-PL205 & 84.25 & 84.18 & 84.18 & 84.32 & \textbf{84.55} & 84.40\tabularnewline
ResNet-PL365 & 87.84 & \textbf{87.91} & 87.84 & 87.80 & 87.76 & 87.69\tabularnewline
\hline
\end{tabular*}
\par\end{centering}
\noindent \centering{}
\end{table}

\subsubsection{Effect of different subgraph combination methods\label{subsec:Effect-of-different}}

\begin{table}
\noindent \begin{centering}
\caption{\label{tab:A-comparison-of-combination}A comparison of different
subgraph combination methods without global information. Classification
accuracy (\%) is reported as evaluation metric. The best result of
each column is marked in bold.}
\par\end{centering}
\noindent \centering{}%
\begin{tabular}{ccc}
\hline
Method & VGG-PL205 & ResNet-PL365\tabularnewline
\hline
Addition & 85.07 & 88.28\tabularnewline
Maximum & 84.93 & 88.13\tabularnewline
Product & \textbf{85.22} & \textbf{88.36}\tabularnewline
\hline
\end{tabular}
\end{table}

We perform a comparison of three different subgraph combination methods,
e.g., element-wise \textit{addition}, \textit{maximum} and \textit{product}.
Table \ref{tab:A-comparison-of-combination} illustrates the results
of LGN without global information on MIT67 dataset. The \textit{product}
combination method outperforms other methods. Compared to \textit{addition}
and \textit{maximum} combination methods, the \textit{product} method
will produce more zero elements in output representations when the
inputs are generated from the ReLU layer. This confirms that the sparsity
of representation is helpful for the improvement of recognition performance.

\begin{table*}
\noindent \begin{centering}
\caption{\label{tab:Ablation-studies}Ablation studies on MIT67. Classification
accuracy (\%) is reported as evaluation metric. The best result of
each column is marked in bold.}
\par\end{centering}
\noindent \centering{}%
\begin{tabular}{cccccccc}
\hline
\multirow{2}{*}{Baseline} & \multirow{2}{*}{Spatial} & \multirow{2}{*}{Similarity} & Global & Node & \multicolumn{3}{c}{Pretrained Model}\tabularnewline
\cline{6-8}
 &  &  & Information & Weak-supervision & VGG-IN & VGG-PL205 & ResNet-PL365\tabularnewline
\hline
$\checkmark$ & - & - & - & - & 74.55 & 81.27 & 86.64\tabularnewline
\hline
- & $\checkmark$ & - & - & - & 77.39 & 84.25 & 87.84\tabularnewline
- & $\checkmark$ & - & - & $\checkmark$ & 78.36 & 84.55 & 87.91\tabularnewline
- & - & $\checkmark$ & - & $\checkmark$ & 78.58 & 84.25 & 87.99\tabularnewline
- & $\checkmark$ & $\checkmark$ & - & $\checkmark$ & 78.73 & 85.22 & \textbf{88.36}\tabularnewline
- & $\checkmark$ & $\checkmark$ & $\checkmark$ & $\checkmark$ & \textbf{79.78} & \textbf{85.37} & 88.06\tabularnewline
\hline
\multicolumn{4}{c}{Improvement Over Baseline} &  & 5.23 & 4.1 & 1.42\tabularnewline
\hline
\end{tabular}
\end{table*}

\subsubsection{Ablation Study}

We conduct detailed ablation studies of our LGN on MIT67 dataset in
Table \ref{tab:Ablation-studies}. We analyze the effect of four components,
two subgraphs, global information, and node weak-supervision mechanism
across three different pretrained models. The $\ell_{2}$ normalized
input representations of local regions are averagely pooled as the
inputs to a linear SVM classifier, and then produce the classification
results as baselines. In Table \ref{tab:Ablation-studies}, the best
results are marked in bold, which show the improvements of up to \textbf{$\boldsymbol{5.23\%}$}
over baselines. When applying node weak-supervision mechanism, LGN
with VGG-IN has a better improvement. It can be attributed to the
worse local representations for scene recognition. Moreover, it can
be observed that the global information is useful for VGG16 pretrained
models, but not for ResNet50 pretrained model. This may be caused
by the better representations of local regions from ResNet-PL365 for
capturing the whole image information. We also validate that the spatial
and similarity subgraphs are both important to boost the performances
and have similar improvements over the baselines. Furthermore, when
combining these two subgraphs, there still have improvements, which
demonstrates that the two subgraphs exist a complementary relation.

\section{Conclusion}

We propose to construct Prototype-agnostic Scene Layout (PaSL) for
each image, and introduce Layout Graph Network (LGN) to explore the
spatial structure of PaSL for scene recognition. The pretrained CNN
models can be used as region detectors to discover discriminative
regions, then form PaSL for each image. To preserve the diverse and
flexible spatial structures of PaSLs, we reform each PaSL as a layout
graph where regions are defined as nodes and two kinds of independent
relations between nodes are encoded as edges. Then, LGN applies graph
convolution on the layout graph to integrate spatial and semantic
similarity relations into image representations. The detailed ablation
experiments demonstrate that LGN has a great ability to capture the
spatial and similarity information in PaSL. With the qualitative and
quantitative analyses, we prove that PaSLs can capture the useful
and discriminative information of the images and have the considerable
generalization capability. Experiments on three widely recognized
datasets, MIT67, SUN397, and Places365, demonstrate that our approach
can achieves superior performances in the setting of a single model
in a single scale, and even obtains state-of-the-art results on MIT67
and SUN397.

In the future, we consider jointly learning scene layout and structural
models, which may bring better optimization results. Another interesting
direction is to explore the multi-scale information from different
convolutional layers to help construct more precise and useful spatial
structures of scene images.

{\small{}\bibliographystyle{IEEEtran}
\bibliography{ms}
}{\small \par}
\end{document}